%% file: main.tex
\documentclass{INTERSPEECH2023}
\usepackage{xcolor}

\interspeechcameraready 

\newcommand{\token}[1]{$\langle\texttt{#1}\rangle$}
\newcommand{\mtoken}[1]{\langle\texttt{#1}\rangle}

\title{Text Injection for Capitalization and Turn-Taking Prediction in Speech Models}
\name{Shaan Bijwadia$^1$, Shuo-yiin Chang$^1$, Weiran Wang$^1$, Zhong Meng$^1$, Hao Zhang$^1$, Tara N. Sainath$^1$}
\address{
  $^1$Google, USA
}
\email{\{shaanb, shuoyiin, weiranwang, haozhang, zhongmeng, tsainath\}@google.com}

\begin{document}

\maketitle

\input{1_abstract}
\input{2_introduction}
\input{3_related}
\input{4_tasks}
\input{5_model}
\input{6_training}
\input{7_experiment}
\input{8_results}

\bibliographystyle{IEEEtran}
\bibliography{refs}

\end{document}

%% file: 1_abstract.tex
\begin{abstract}
Text injection for automatic speech recognition (ASR), wherein unpaired text-only data is used to supplement paired audio-text data, has shown promising improvements for word error rate. This study examines the use of text injection for auxiliary tasks, which are the non-ASR tasks often performed by an E2E model. In this work, we use joint end-to-end and internal language model training (JEIT) as our text injection algorithm to train an ASR model which performs two auxiliary tasks. The first is capitalization, which is a de-normalization task. The second is turn-taking prediction, which attempts to identify whether a user has completed their conversation turn in a digital assistant interaction. We show results demonstrating that our text injection method boosts capitalization performance for long-tail data, and improves turn-taking detection recall.
\end{abstract}

%% file: 2_introduction.tex
\section{Introduction}
\label{sec:introduction}

Automatic speech recognition (ASR) has long been an integral part of important technologies, including voice dictation, digital assistants, and video captioning \cite{Schalkwyk2010}. While ASR systems are typically evaluated based on word error rate (WER), this is not the only metric of concern in production applications; several ``auxiliary tasks" must be integrated with the ASR task in a full system. These tasks may include: capitalization and punctuation, which improves readability; voice activity detection (VAD) and end-of-query (EOQ) detection, which are important for implementing low-latency systems; and natural conversation understanding, which involves predicting the cadence and turn-taking aspects of an ongoing conversation. In this study, we focus on improving the quality of such auxiliary tasks in an end-to-end (E2E) ASR setting via text injection.

We build on two recent capabilities for speech models. First is the E2E integration of auxiliary tasks with the ASR task into a single model. In the past, auxiliary tasks were usually performed by separate models downstream of ASR \cite{chang2017endpoint, Beaufays2013Cap, Nguyen2019FastAA, Batista2011RecoveringCA}. Recent work has successfully explored integrating auxiliary tasks, such as endpointing \cite{Bijwadia2023Endpointer, Shuoyiin19, maas2018}, capitalization \cite{Wang2023Multi}, natural conversation understanding \cite{Chang2022TurnTakingPF}, and speaker diarization \cite{Shafey2019JointSR} into the same model as ASR prediction. E2E integration of ASR and auxiliary tasks has a key drawback, however. When folded into an E2E ASR model, pure text-to-text tasks (such as capitalization) can no longer be trained on plentiful text-only data (i.e., text data with no associated audio); instead, their training examples will be limited to the transcripts available in paired audio-text labeled data. This puts E2E methods at a disadvantage, since text-only data is generally more plentiful and easier to obtain than labeled audio data, and can be used to more easily expose the model to rare words and other long-tail phenomena which may be difficult to collect in labeled audio form \cite{Huang2022SentenceSelectLL}.

The second capability enabling the current study is the use of ``text injection" as a means of improving ASR quality \cite{Thomas2022IntegratingTI}. An ASR model's internal language model (ILM) is the notional part of the network that predicts the next token given the previous token history, independent of audio input. Though it is usually infeasible to exactly separate the influence of audio input from previous token predictions, several methods have been developed to estimate ILM scores \cite{Variani20, Meng2020InternalLM}. Text-only data can then be used to refine the ILM capabilities of the ASR network \cite{Sainath2022JOISTAJ, Chen2022Maestro}.

In this work, we propose a method to utilize text injection techniques for improving auxiliary task performance in an E2E ASR model. Doing so allows auxiliary tasks to access the multi-task learning benefits of co-training with ASR while still including rich text-only data in their training corpora. We focus our study on two tasks: capitalization and conversational turn-taking prediction. The former is a strongly text-based task, since capitalization is merely a form of de-normalization from spoken to written domain, and capitalized words are not pronounced differently. The latter task clearly involves combining linguistic and acoustic understanding --- the prosody of the input speech as well as the semantics of the current recognition are both informative for predicting whether a pause is only momentary or if the user has finished speaking. We integrate these tasks into a production-ready model, streaming E2E RNN-T ASR model \cite{Graves12, sainath2020streaming}. We show results demonstrating that text injection can meaningfully improve auxiliary task performance, particularly in long-tail settings.


%% file: 3_related.tex
\section{Related Work}
\label{sec:related}

While auxiliary tasks are usually performed by separate models from ASR \cite{Rei2020AutomaticTO, Sunkara2020RobustPO}, E2E approaches to auxiliary task modeling have been recently popular for production-grade systems. Joint training of ASR with endpointing \cite{Shuoyiin19}, capitalization \cite{Wang2023Multi, Pahuja2017JointLO}, intended query detection \cite{Chang2022StreamingIQ, Mallidi2018DevicedirectedUD}, sentence segmentation \cite{Huang2022E2ESJ}, and more, have been explored. Our work builds most closely on Wang\nobreakspace et\nobreakspace al.\cite{Wang2023Multi}, who co-train ASR, capitalization, and turn-taking prediction by building multiple parallel label sequences. To our knowledge, this is the first attempt to refine auxiliary tasks in an E2E ASR model using text-only data.

There has long been interest in utilizing unpaired text data for the ASR task. Several approaches to LM fusion, the use of an external LM to improve ASR recognition quality, have been proposed \cite{Anjuli18}. These methods have the drawback of increasing total parameter count (due to the size of the LM model), and computation cost during inference. Text injection \cite{Thomas2022IntegratingTI} solves these issues by using LM-style unpaired text data to train the ASR model itself. Some methods focus on fine-tuning an existing ASR model trained on audio-text data; ILM adaptation of the ASR decoder has been shown to work well \cite{Meng2021InternalLM, Bataev2023TextonlyDA, Meng2022MHAT}. The text injection method we employ here is joint end-to-end and ILM training (JEIT), which was introduced by Meng et al \cite{Meng2023Jeit}. We choose JEIT as our method due to its lightweight nature; its primary focus on refining the ASR decoder makes comparison to standard methods straightforward, since the behavior of the audio encoder is preserved. Other methods inject text data directly into the encoder, with fixed and learned duration models to align text and audio sequences \cite{Sainath2022JOISTAJ, Chen2022Maestro}. All of the above works focus on improving ASR quality, both for standard and long-tail data; to the best of our knowledge, adapting these techniques for auxiliary tasks is a novel contribution to the literature.

%% file: 4_tasks.tex
\section{Auxiliary Tasks}
\label{sec:tasks}

\subsection{Capitalization}
\vspace*{-0.5ex}

Capitalization is the process of restoring the correct case (uppercase or lowercase) of noisy text. Notably, capitalization is specific to the written domain, and has no marker in spoken speech. This task is important for maintaining readability in ASR output, especially for long-form captioning cases.

\subsection{Conversational turn-taking}
\vspace*{-0.5ex}

Turn-taking is an active area of research for E2E speech modeling \cite{Chang2022TurnTakingPF, Liu2017TurnTakingEM}. While humans typically adjust their speech when interacting with voice assistants \cite{Liu2017TurnTakingEM}, natural human speech patterns during conversation are often filled with natural disfluencies. For digital assistant products, it is desirable that voice assistants have the ability to predict when the speaker is expecting a response, versus when they merely pause with the intention to resume speaking. We model this phenomenon similar to Chang et al. \cite{Chang2022TurnTakingPF}, who classify pauses in speech as being within a complete thought, or after having a finished complete thought. That is, when a user stops speaking, the model should predict whether they will continue speaking after a brief pause or whether a system response is expected. Because the active region of interest is pauses in the audio, we refer to this task in this paper as ``pause prediction."

\begin{figure}[t]
  \centering
  \includegraphics[width=0.8\linewidth]{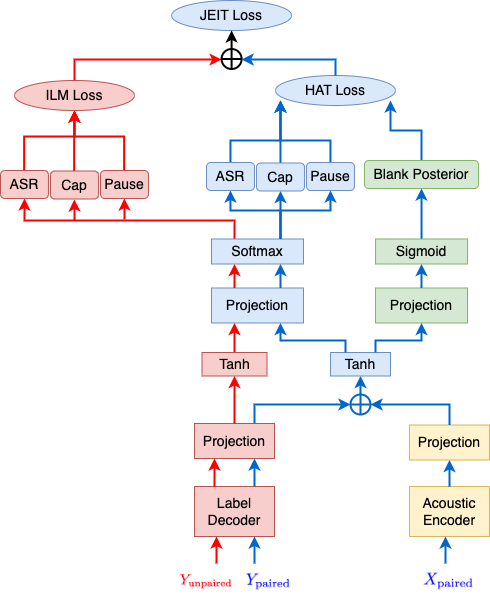}
  \vspace{-0.15in}
  \caption{Model diagram for JEIT training. The blue arrows denote the data flow for paired audio-text data. The red arrows denote the path that unpaired text data takes through the network. Baseline experiments are trained using only the blue paths, while the proposed system is trained using both.}
  \label{fig:architecture}
  \vspace{-0.2in}
\end{figure}

%% file: 5_model.tex
\section{Model}
\label{sec:model}

\begin{figure*}
    \centering
    \begin{tabular}{|c|l|} \hline
    Initial Transcript &\hspace{0.4in} driving time to san francisco \\[.2ex] \hline
    Capitalization &\hspace{0.4in} Driving time to San Francisco \\ \hline
    Tokenization &\hspace{0.4in} \token{cap} \_driving \_time \_to \token{pause} \token{cap} \_san \token{cap} \_fran cisco \token{eos} \\ [.2ex] \hline
    Label Factorization
    &$Y_{\text{ASR}}:$\quad \_driving \hspace{0.35in} \_time \hspace{0.48in} \_to \hspace{0.45in} \_san \hspace{0.52in} \_fran \hspace{0.51in} cisco \\
    &$Y_{\text{Cap}}:$\quad \token{cap} \hspace{0.42in} \token{non-cap} \hspace{0.11in} \token{non-cap} \hspace{0.0in} \token{cap} \hspace{0.42in} \token{cap} \hspace{0.42in} \token{non-cap} \\
    &$ Y_{\text{Pause}}:$\hspace{0.07in}\token{non-pause} \token{non-pause} \token{pause} \hspace{0.15in} \token{non-pause} \token{non-pause} \token{eos} \\ \hline
    \end{tabular}
    \caption{Data preparation for auxiliary tasks. Wordpieces that begin with \_ denote word boundaries. In this example, we assume that the speaker takes a verbal pause as follows: "Driving time to... San Francisco," to illustrate the \token{pause} logic.}
    \label{fig:labels}
    \vspace{-0.2in}
\end{figure*}

\subsection{Multi-output HAT decoder}
\label{subsec:decoder}
\vspace*{-0.5ex}

HAT is a decoder structure for RNN-T in which the \token{blank} probability is computed separately from next token prediction, facilitating more accurate ILM estimation \cite{Variani20}. Wang et al. \cite{Wang2023Multi} propose a variant of HAT decoder which introduces multiple joint networks, one for each task (in our case, these are ASR, capitalization, and pause prediction). All of the parallel joint networks are conditioned on features from both the prediction network and audio encoders.

The model is trained using an RNN-T objective \cite{Graves12}, where at each timestep the model may choose to emit a wordpiece token, or to insert a special token \token{blank} which indicates non-emission. Formally, let $X$ be the input utterance and $Y$ be the label sequence. The ASR output space $\mathcal{Y_{\text{ASR}}}$ consists of $\{y^{0}=\mtoken{blank}, y^{1}, y^{2}, ...\}$.. Let $T=|X|$ be the number of input audio frames and $U=|Y|$ be the length of the transcript. The acoustic encoder produces $f(X)=[f_0,...,f_{T-1}]$, $f_t \in \mathcal{R}^{D_a}$, and the prediction network produces $g(X)=[g_0,...,g_{U-1}]$, $g_u \in \mathcal{R}^{D_p}$. As in the original HAT implementation, the joint network fuses $f_t$ and $g_u$ with a ``project and sum" operation to produce a hidden representation $h_{t,u}$, which is then passed through a non-linear activation and a final linear layer to produce $s_{t,u}$:
\begin{align}
    h_{t,u} = P \cdot f_t + Q \cdot g_u + b_h \quad \in \mathcal{R}^{D_h} \\
    s_{t,u} = A \cdot \text{tanh} (h_{t,u}) + b_s \quad \in \mathcal{R}^V.
\end{align}
where $P$, $Q$, and $A$ are learned weight matrices with dimensions determined by $D_a$, $D_p$, $D_h$, and $V$ is the size of the vocabulary. As this is a HAT model, the 0-th logit of $s_{t,u}$ is used individually to compute the probability of emission $b_{t,u}$:
\begin{align}
    b_{t, u} := P_{t,u}(\mtoken{blank} | f_{0:t}, g_{0:u}) = \sigma (s_{t,u} [0])
\end{align}
where $\sigma (x) = 1 / (1+\exp(-x))$ is the sigmoid activation. Probabilities over the ASR tokens are computed by feeding all remaining logits to a softmax function. The probability of each ASR token $y_v$ in the vocabulary is:
\begin{align}
    \hat{y}_{v; t,u} &= P_{t,u}(\hat{y}_{v} | f_{0:t}, g_{0:u}) \notag\\
    &= \text{softmax}(s_{t,u}[1:])[v-1]
\end{align}
Thus the predicted probability distribution over all output tokens is the emission probability, followed by the probabilities of each token given emission:
\begin{align}
    \hat{y}_{t,u} = [&b_{t,u}, \quad (1-b_{t,u})\cdot \hat{y}_{0; t, u}, \notag\\
    &\quad ...\ , \quad (1-b_{t,u})\cdot \hat{y}_{V-1; t, u}]
\end{align}

Thus far we have referred to the mechanism above in terms of ASR prediction. Capitalization and pause predictions are made in the exact same way, where each task independently computes Eqs. (1) and (2) based on the shared representations $f_t$ and $g_u$ (note that each auxiliary task is exposed to the label history of the ASR output, not its own prediction history). 

Since capitalization tokens must be strictly aligned with ASR tokens, the capitalization posterior borrows the blank logit from the ASR prediction. Thus, a capitalization token will only be emitted when an ASR token is emitted as well. Capitalization has output space $\mathcal{Y}_{\text{Cap}} = \{\mtoken{cap}, \mtoken{non-cap}\}$ and its posterior is:
\begin{align}
    \hat{y}^{\text{Cap}}_{t,u} = [&b^{\text{ASR}}_{t,u}, \quad (1-b^{\text{ASR}}_{t,u})\cdot P_{t,u}(\mtoken{cap}), \notag \\
    &(1-b^{\text{ASR}}_{t,u})\cdot P_{t,u}(\mtoken{non-cap})]
\end{align}

At inference time, we estimate $P(\mtoken{cap})$ every time an ASR token is emitted and predict a capitalization if it is above a threshold (in this work, we use 0.5).

Pause tokens do not need to be strictly aligned with the ASR transcript prediction, since they are likely to be predicted during non-speech periods in the audio during inference, so the turn-taking sequence has its own blank posterior. The pause prediction output space is $\mathcal{Y}_{\text{Pause}} = \{\mtoken{blank}, \mtoken{non-pause}, \mtoken{pause}, \mtoken{eos}\}$ and its posterior is computed in the same way as Eq. (5).

%% file: 6_training.tex
\section{Training}
\label{sec:training}

\subsection{JEIT}
\label{subsec:jeit}
\vspace*{-0.5ex}

Joint end-to-end model and ILM training (JEIT) was proposed by Meng et al. \cite{Meng2023Jeit} as a way to train an RNN-T ASR model on paired audio-text data while simultaneously training the HAT decoder ILM on text-only data. For paired dataset $\mathcal{D}_{\text{paired}}$, training is conducted in the usual way; the model is given the audio sequence as input and predicts $P_{\text{E2E}}(Y | X)$. This is converted to a loss $\mathcal{L}^{\text{ASR}}_{\text{E2E}}$ via the RNN-T objective \cite{Graves12}.

The text-only dataset $\mathcal{D}_{\text{unpaired}}$ contains transcripts with capitalization and pause annotations (see \S \ref{subsec:labels}). Similar to HAT ILM adaptation (ILMA) \cite{Meng2021InternalLM}, we feed the transcript as the previous token history to the prediction network, and mock the encoder output with vectors full of zeros: $\forall_{t\in 0:T}: f_t = \bold{0}$. Since the audio sequence does not exist, we simply ignore the blank posterior, and the predicted next token probabilities are given directly by the softmax output in Eq. (4). With previous token history as input and next token probabilities as output, this allows us to estimate $P_\text{ILM}(y_t : y_{0:t-1})$. ILM loss is defined as the negative log probability of each label token given the label sequence prefix:
\begin{align}
    \mathcal{L}^{\text{ASR}}_{\text{ILM}} = - \sum^{U}_{u=1} \log P(y^{\text{ASR}}_{u} | \hat{y}^{\text{ASR}}_{0:u-1})
\end{align}

The losses $\mathcal{L}_{\text{E2E}}$ and $\mathcal{L}_{\text{ILM}}$ are averaged over their respective datasets $\mathcal{D}_{\text{paired}}$ and $\mathcal{D}_{\text{unpaired}}$, then combined in a weighted average to obtain the total JEIT loss:
\begin{align}
    \mathcal{L}^{\text{ASR}}_{\text{JEIT}}(\mathcal{D}_{\text{paired}}, \mathcal{D}_{\text{unpaired}}) &= \notag\\ \mathcal{L}^{\text{ASR}}_{\text{E2E}}(\mathcal{D}_{\text{paired}}) &+ \beta \mathcal{L}^{\text{ASR}}_{\text{ILM}}(\mathcal{D}_{\text{unpaired}})
\end{align}
where $\beta$ is a hyperparameter controlling the weight given to ILM training (in this work, we use $\beta=0.2$ to match Meng\nobreakspace et\nobreakspace al.'s original study).

Adapting JEIT to include auxiliary tasks is straightforward. As described in \S \ref{subsec:decoder}, each auxiliary task makes a sequence prediction $Y_{\text{Aux}}$ based on the predicted ASR sequence $Y_{\text{ASR}}$. Thus, each auxiliary task predicts $P_{\text{E2E}}(Y_{\text{Aux}}|\hat{Y}_{\text{ASR}}; X)$ to produce $\mathcal{L}^{\text{Aux}}_{\text{E2E}}$. Similarly, the ILM loss is
\begin{align}
    \mathcal{L}^{\text{Aux}}_{\text{ILM}} = - \sum^{U}_{u=1} \log P(y^{\text{Aux}}_{u} | \hat{y}^{\text{ASR}}_{0:u-1})
\end{align}
The full JEIT loss for each task is defined in the same way as Eq. (8). Total loss is a linear combination of all tasks: (datasets omitted for clarity):
\begin{align}
    \mathcal{L}^{\text{Total}}_{\text{JEIT}} =& \quad \mathcal{L}^{\text{ASR}}_{\text{E2E}} + \beta
    \mathcal{L}^{\text{ASR}}_{\text{ILM}}
    &&+\alpha_{\text{Cap}}(\mathcal{L}^{\text{Cap}}_{\text{E2E}} + \beta \mathcal{L}^{\text{Cap}}_{\text{ILM}}) \notag\\
    &&&+\alpha_{\text{Pause}}(\mathcal{L}^{\text{Pause}}_{\text{E2E}} + \beta \mathcal{L}^{\text{Pause}}_{\text{ILM}})
\end{align}
where each $\alpha$ is a loss weight for the corresponding task. Matching Wang's original study, we use $\alpha_{\text{Cap}} = 0.1$ and $\alpha_{\text{Pause}} = 0.3$. Figure \ref{fig:architecture} shows the data flow for paired and unpaired data through the ASR model.

\subsection{Transcript annotation}
\label{subsec:labels}
\vspace*{-0.5ex}

While a small amount of our paired training corpus is hand-labeled and capitalized, most of our paired data and all of our unpaired text data have lowercase transcripts. For the lowercase transcripts, we use a text-based truecasing RNN teacher model similar to \cite{Zhang2022Capitalization} to produce capitalization predictions.

Producing pause prediction labels requires different approaches for paired and unpaired data. For paired audio-text data, we use the approach taken by Chang et al. \cite{Chang2022TurnTakingPF}, which uses heuristics based on a forced alignment \cite{moreno1998recursive} to insert pause tokens into the transcript. There are two such tokens: \token{pause} denotes a brief stop by the speaker in the middle of a full thought, and \token{eos} (end of sentence) is inserted at the end of the full thought, i.e. a full conversational turn.

For unpaired text-only data, the above strategy is impossible, since we do not have access to the associated audio. Instead, we rely on the fact that our text-only data comes from short-query sources (see \S \ref{subsec:data}). We simply append the \token{eos} token to the end of the transcript.

\subsection{Multi-task label structure}
\label{subsec:factorization}
\vspace*{-0.5ex}

A common approach to transcript labeling for auxiliary tasks would be to embed special tokens corresponding to each task in the transcript itself \cite{Shuoyiin19}. However, this is not ideal for inference, since the extra tokens must be expanded in-line with the ASR tokens; if predictions on competing beams differ only in their special tokens, lattice diversity is reduced because the ASR prediction would be identical. To solve for this, we follow Wang\nobreakspace et\nobreakspace al. \cite{Wang2023Multi}, factorizing the auxiliary task tokens into parallel sequences of equal length, one for each task. The ASR task is trained on the lowercase transcript sequence, segmented into wordpieces. The capitalization sequence is defined as follows: each token is either \token{cap} (capitalized) or \token{non-cap} (not capitalized), based on the corresponding wordpiece in the ASR transcript. Similarly, the turn-prediction sequence is populated with \token{pause} and \token{eos} tokens corresponding to the wordpieces immediately preceding the corresponding predicted pauses in the transcript. All other token slots are filled with \token{non-pause}. The successive steps of label generation are shown in Figure \ref{fig:labels}.

%% file: 7_experiment.tex
\section{Experimental Details}
\label{sec:experiment}


\begin{table}[t]
    \centering
    \caption{Capitalization. We report word error rate (WER (\%)) and uppercase error rate (UER (\%)) on a representative (``head") voice dictation dataset. We also report UER on a dataset containing rare words (``tail").}
    \vspace{-0.1in}
    \label{tab:cap-results}
    \begin{tabular}{c|c|c|c|c}
    \toprule
        \textbf{Exp.} & \textbf{Method} & \textbf{WER} & \textbf{Head} & \textbf{Tail}  \\
        & & & \textbf{UER} & \textbf{UER} \\ \hline
        B1 & Paired Data Only &  3.9 & 24.3 & 46.0 \\ \hline
        E1 & JEIT (Proposed) & 3.9 & 24.7 & 45.1 \\
    \bottomrule
    \end{tabular}
    \vspace*{-1ex}
\end{table}

\begin{table}[t]

    \centering
    \caption{Sample capitalization improvements. For anonymity, some transcript words are substituted with equivalents, while preserving the capitalization dynamics.}
    \vspace{-0.1in}
    \label{tab:cap-examples}
    \resizebox{\columnwidth}{!}{
    \begin{tabular}{c|c|c}
    \toprule
        \textbf{Ground Truth} & \textbf{Hypothesis} & \textbf{Exp.} \\ \hline
        Matheus Nicolau UFC fighter &  Matheus nicolau UFC fighter & B1 \\
        & Matheus Nicolau UFC fighter & E1 \\ \hline
        Smoketown Brewing Company &  smoketown Brewing Company & B1 \\
        & Smoketown Brewing Company & E1 \\ \hline
        Play Maldita Vecindad &  play Maldita vecindad & B1 \\
        & play Maldita Vecindad & E1 \\
    \bottomrule
    \end{tabular}}
    \vspace*{-1ex}
\end{table}

\begin{table}[t]
    \centering
    \caption{Pause prediction. We report precision and recall for the \token{eos} token on a conversation-style test set.}
    \vspace{-0.1in}
    \label{tab:pause-results}
    \begin{tabular}{c|c|c|c}
    \toprule
        \textbf{Exp.} & \textbf{Method} & \multicolumn{2}{c}{\token{eos} Prediction} \\ \cline{3-4}
        & & \textbf{Precision} & \textbf{Recall}  \\ \hline
        B1 & Paired Data Only & 72.16 & 89.56 \\ \hline
        E1 & JEIT (Proposed) & 71.12 & 92.94 \\
    \bottomrule
    \end{tabular}
    \vspace*{-1ex}
\end{table}

\subsection{Model architecture}
\label{subsec:architecture}
\vspace*{-0.5ex}

We use a 128-dimensional log-mel feature frontend computed on 32ms windows with a 10ms stride. We stack four consecutive frames together and sub-sambled by a factor of 3, resulting in 512-dim features at a 30ms framerate. This vector is then concatenated with a 16-dim one-hot domain ID vector \cite{Arun19}.
As our ASR backbone we use a 2-pass cascaded encoder model \cite{arun21cascade}. The first encoder consists of 7 conformer layers \cite{gulati2020conformer} with causal convolution and left-context attention. The second encoder consists of 10 conformer layers with a 900ms lookahead. Each conformer layer uses 512-dim 8-head self-attention and a kernel size of 15, and the final layer emits $D_a = 384$-dim encodings. The prediction network of each decoder is a $V^2$ embedding lookup table, which computes $D_p = 640$-dim features based on embeddings of the previous two wordpiece tokens. Each joint network has hidden dimension $D_h = 384$, and predictions are made over a vocabulary of $V=4096$ wordpieces. For evaluation, we report only 2nd pass WER.

In total, our model has $\sim$160M params. It is implemented in Tensorflow using the Lingvo toolkit, and is trained on proprietary specialized hardware for 500k steps using batch size 4096 for paired and unpaired data.

\subsection{Data}
\label{subsec:data}
\vspace*{-0.5ex}

\subsubsection{Paired  training data}
\vspace*{-0.5ex}
Our training set of audio-text pairs consists of a dataset of ~650 million English multi-domain examples, drawn from search, dictation, online video, and telephony domains. A small subset of these utterances are anonymized and hand-transcribed, and the rest are pseudo-labeled by a 600M parameter bidirectional teacher model. To increase model robustness, we apply simulated noise to utterances, as well as SpecAug \cite{Park2019}.

\subsubsection{Unpaired training data}
\vspace*{-0.5ex}
Our text-only data selection pipeline is designed in the style of Sentence-Select by Huang\nobreakspace et \nobreakspace al \cite{Huang2022SentenceSelectLL}. Text query data ($\sim$ 100B utterances) is collected from web search, maps search, app store search, and online video search domains. This data is filtered for rare words and contrastive filtering based on perplexity is applied. Because the data is selected to include rare words, we expect improvements at the tails of the evaluation distribution.

\subsubsection{Evaluation Data}
\vspace*{-0.5ex}
WER is reported on $\sim$\nobreakspace17k utterances representative of real-world voice dictation traffic. Ground truth transcript and auxiliary task annotations are obtained via human labeling. We also report uppercase error rate (UER) on this set, which is calculated by removing all lowercase letters from the ground truth label and the predicted transcript and computing standard WER with upper case letters as words. Since our text-only data focuses on long-tail traffic, we also report UER on a set of $\sim$300 utterances with transcripts containing rare words.


For pause prediction, we use a testset of $\sim$2500 utterances containing hesitations and multiple consecutive commands. Pauses in the audio are hand-annotated as continuation pauses or final pauses. The metrics reported are average precision and recall of the \token{eos} token.

%% file: 8_results.tex
\section{Results}

We evaluate the proposed method (E1) against a baseline (B1) which uses an identical model but is trained on paired data only (Table \ref{tab:cap-results}). On the large voice search test set on which it is evaluated, WER does not change, while UER regresses slightly on the voice dictation dataset (1.6\% relative). For long tail data, UER improves by a relative 2.0\%. Table \ref{tab:cap-examples} shows example transcripts demonstrating our proposed method's better capability at recognizing capitalized named entities.

Pause detection recall improves by 3.7\% (relative), while precision is reduced slightly, by 1.4\% (relative) (Table \ref{tab:pause-results}). This matches the intuition that our text-injection method biases the model towards \token{eos}, since the unpaired text data is only augmented with \token{eos} at the end of short form transcripts. However, the improvement in recall is larger than the change in precision, and in a production setting, hyperparameters may be tuned to balance the two metrics differently. These results show that augmenting the training data of an ASR model with unpaired text data using JEIT can be used to meaningfully improve pause prediction performance, without regressing word-error rate.

These results show that augmenting the training data of an ASR model with unpaired text data meaningfully impacts auxiliary task performance. In our case, we use long-tail, shortform text data to improve capitalization performance for rare words and turn-taking prediction recall. We recommend that future work extend this technique to other text injection methods, and explore the use of text injection for other auxiliary tasks.